# Early fault detection with multi-target neural networks

Angela Meyer [1][0000-0003-4120-3827]

[1] Bern University of Applied Sciences, 2501 Biel, Switzerland
`angela.meyer@bfh.ch`

**Abstract.** Wind power is seeing a strong growth around the world. At the same time, shrinking profit margins in the energy markets let wind farm managers explore options for cost reductions in the turbine operation and maintenance. Sensor-based condition monitoring facilitates remote diagnostics of turbine subsystems, enabling faster responses when unforeseen maintenance is required. Condition monitoring with data from the turbines' supervisory control and data acquisition (SCADA) systems was proposed and SCADA-based fault detection and diagnosis approaches introduced based on single-task normal operation models of turbine state variables. As the number of SCADA channels has grown strongly, thousands of independent single-target models are in place today for monitoring a single turbine. Multi-target learning was recently proposed to limit the number of models. This study applied multi-target neural networks to the task of early fault detection in drive-train components. The accuracy and delay of detecting gear bearing faults were compared to state-of-the-art single-target approaches. We found that multi-target multi-layer perceptrons (MLPs) detected faults at least as early and in many cases earlier than single-target MLPs. The multi-target MLPs could detect faults up to several days earlier than the single-target models. This can deliver a significant advantage in the planning and performance of maintenance work. At the same time, the multi-target MLPs achieved the same level of prediction stability.

## 1 Introduction

The global wind power capacity is growing strongly with a total installed volume of 651 GW in 2019 and an increase of 76 GW in 2020 [1]. The newly installed wind turbines are getting larger and increasingly more complex. At the same time, the operating cost of wind farms still makes up a major fraction, approximately 30%, of their lifetime cost [2]. Major faults can result in days and even weeks of downtime [3,4]. Therefore, they can substantially reduce the owner's return on investment and pose a considerable economic risk. As a result, many operators want to closely monitor the health state of their turbines in order to be alerted as early as possible of any developing technical problems and to prevent any major damage and downtime. To this end, an automated condition monitoring of wind turbine subsystems provides an essential pre-requisite for informed operational decision making and fast responses in case of unforeseen maintenance needs [5-6].

Data-driven automated monitoring methods have been proposed, amongst others, based on sensor data logged in the turbines' supervisory control and data acquisition (SCADA) systems [7-10]. Temperature can be an important indicator of different types of developing machine problems such as mechanical faults which can give rise to excessive friction generating heat. Therefore, a major focus of the proposed SCADA-based condition monitoring approaches is the temperature-based detection of developing faults in the wind turbine subsystems based on models of the turbine's normal operation behaviour in the absence of operational faults [11-22]. The present study focuses on the gear bearing temperature as an indicator of developing gearbox faults.

The goal of this study is to assess the potential of multi-target regression models for the automated SCADA-based fault detection. Specifically, this work has investigated and compared the delays in detecting gear bearing faults using single-target versus multi-target models of the turbines' normal operation. Moreover, the stability of the alarm signal after the first detection of a developing fault is being assessed.

The remainder of this paper is structured as follows. Section 2 provides a brief overview of previous work in this field. Section 3 describes the data sources and the training and testing of the multi- and the single target regression models. The analysis and results are discussed in section 4. Conclusions and possible future work are proposed in section 5.

## 2 Related Work

Normal behaviour modelling has become an established technique in wind turbine condition monitoring and fault detection [7-8]. Normal behaviour models characterize the machine state during normal operation in the absence of faults. They have been in use for monitoring the health state of turbine subsystems such as the gearbox [23,24] and the generator [25]. Normal behaviour models have also successfully been employed for monitoring the active power generation [26-27,15]. We refer to [8] for a comprehensive review of SCADA-based condition monitoring and normal behaviour models of wind turbines.

Multi-target machine learning models [28-32] are regression or classification models which predict multiple target variables simultaneously. It has been demonstrated in other fields that multi-target models hold the potential to enable an increased prediction accuracy compared to single-target models and are less susceptible to overfitting the training data [29, 33-34]. In the field of wind turbine monitoring, we have recently introduced multi-target regression models for simultaneously monitoring the growing number of SCADA channels, and we demonstrated that they can reduce the effort of SCADA-based normal behaviour monitoring in wind turbine condition monitoring [35].

# 3      Data and Methods

Condition monitoring data from the SCADA system of three commercial onshore turbines was analyzed in this work. The turbines are variable-speed three-bladed horizontal axis systems with pitch regulation from an onshore wind farm. Their rated power was 3.3 MW, and they operated with a 3-stage planetary/helical gearbox. The turbines' rotors were 112 m in diameter with the hub located at 84 m height above ground. The turbines' cut-in, rated and cut-out wind speeds were specified at 3m/s, 13 m/s and 25 m/s, respectively.

In this study, fourteen months of ten-minute mean SCADA signals served to train and test the models specified below. The data were anonymized to maintain the privacy of the wind farm operator. We report the results for one of the wind turbines. It was randomly selected and the results were not affected by the choice of turbine. We focus on monitoring the gear bearing condition based on the temperature of the gear bearing. The temperature is an important SCADA-based indicator of incipient fault processes in gearbox components [8,19, 20, 22-23]. In the present study, the condition of the gear bearing has been monitored based on two normal operation models of the bearing temperature. Wind speed $v_{wind}$, wind direction $\alpha_{wind}$ and air temperature $T_{air}$ constitute the models' input variables which were provided as ten-minute averages of measurements from nacelle-mounted anemometers and thermometers. The input variables were selected due to their relevance for explaining and predicting the behaviour of the target variables.

A multi-target fully connected feedforward neural network was designed to predict the gear bearing temperature $T_{gear}$ along with the hydraulic oil temperature $T_{oil}$ and the transformer winding temperature $T_{tr}$ from the input variables at high accuracy, $T_{gear}$, $T_{oil}$, $T_{tr} \sim v_{wind} + \alpha_{wind} + T_{air}$. The single-target model estimates the gear bearing temperature only, $T_{gear} \sim v_{wind} + \alpha_{wind} + T_{air}$. In addition, the two fully connected feedforward neural networks (multi-layer perceptrons, MLP) were trained and tested to assess the normal operating behaviour of the gear bearing based on the provided SCADA data. The model architectures were developed to obtain a high predictive accuracy on the training set without overfitting the training data. In this process, the number of neurons and weights to be trained was increased only if this resulted in higher predictive accuracy. The resulting model architectures are detailed in Table 1.

In this work, our goal is to systematically assess the ability of multi-target neural networks to detect developing faults that result in rising component temperatures. We demonstrate this approach by the example of the gear bearing temperature. However, it is equally applicable to faults in other subsystems and other components that result in elevated SCADA-logged temperatures.

A major challenge in data-driven fault detection and isolation is the scarcity of actually observed fault instances. We addressed this point by combining gear bearing temperature measurements with a multitude of synthetic temperature trends in order to mimic the bearing temperature rise induced by a developing fault. To this end, a synthetic temperature trend was overlaid on the normalized gear bearing temperature signal. One of ten different linear temperature trends was added to the normalized bearing temperature. Temperature trends with integer slopes in the range of 1 to 10 were used to simulate slowly and fast evolving fault processes. The temperature trend onset time was randomly sampled from a two-week time window in months 12 and 13 of the 14-months observation period. Fifty different onset times have been randomly sampled from the two-week window for each of the ten temperature slopes. This ensured that the results did not depend on the choice of the trend onset time.

**Table 1.** The architectures of the multi-layer perceptrons.

| Model | Architecture |
|---|---|
| Multi-layer perceptron (MLP) with three target variables | Two dense hidden layers with 4 neurons in the first layer and 19 neurons in the second layer, batch normalization, and a 3-neuron output layer. Dropout was applied at a rate of 10% to avoid overfitting. |
| Single-target multi-layer perceptron | Three dense hidden layers with 4 neurons each in the first and second layers and 5 neurons in the third layer, batch normalization and a single node in the output layer. Dropout rate of 10%. |

## 4 Results and Discussion

Two common alarm criteria [7] were applied using the residuals of the gear bearing temperature. The residuals were computed as the difference of the observed temperature of the gear bearing and its temperature predicted by the normal operation behaviour models of Table 1. Figure 1 illustrates the residuals and alarms. According to the first criterion, an alarm was raised when the $99.9^{th}$ percentile of the residuals distribution was exceeded for more than 8 hours in the past 24 hours. On the other hand, the second criterion triggered an alarm whenever the rolling mean of the residuals computed over the past 8 hours exceeded the $99.9^{th}$ percentile of the residuals distribution. The fault detection capabilities of the multi-target and single-target normal behaviour models were compared with regard to the delay of detecting the induced faults. Moreover, the stability of the triggered alarms was assessed following the detection. The stability is computed as the fraction of true positive alarms after the first detection.

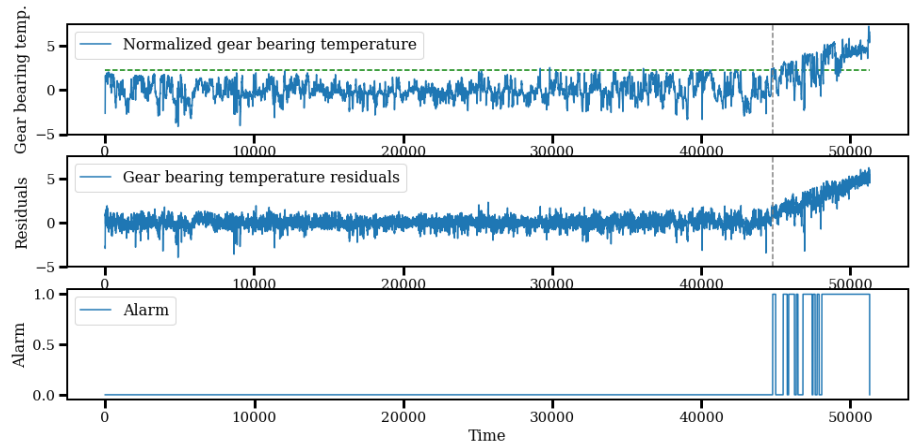

**Figure 1.** Normalized gear bearing temperature and residuals. An alarm was raised when the 99.9[th] percentile of the residuals distribution was exceeded for more than 8 hours during the past 24 hours (alarm criterion 1). The dashed line indicates the first alarm event. The temperature units are dimensionless due to normalization.

Figure 2 shows the resulting alarms for a fault instance with fast rising bearing temperature and a randomly sampled onset time. The false positive detection rate is zero for both criteria but there are false negatives after the first alarm detection by both alarm criteria. Following the first detection, the alarms are initially unstable and stabilize within three to seven days after the first alarm was triggered, as shown in Figure 2. We found that the alarm signal stabilized faster in the case of the first criterion, while the second criterion generated a less stable signal that required a week to stabilize in the case shown in Figure 2. This alarm instability was caused by the signal-to-noise ratio which was relatively small immediately after the first detection and increased over time along with the overlaid increasing temperature.

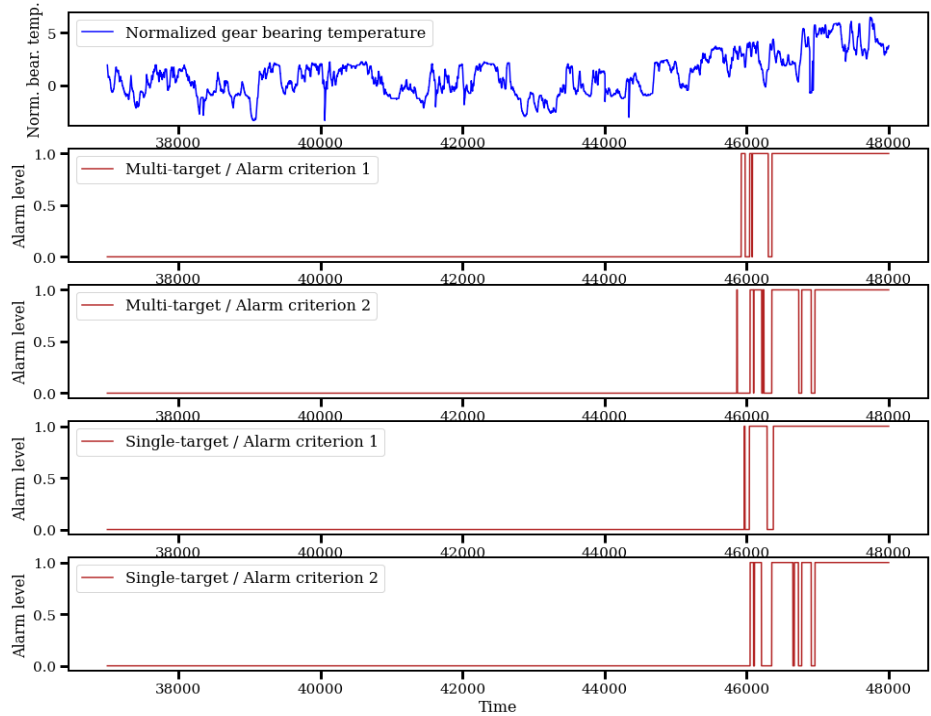

**Figure 2.** A fast evolving gear bearing fault with trend slope 10 is shown at one of the 50 randomly sampled onset times. The gear bearing temperature starts to rise near time step 44000 in this case. The alarm criteria are compared based on the multi-target and the single-target normal behaviour models of the gear bearing temperature. Alarm criterion 2 detects the trend earlier but is less stable than alarm criterion 1.

The detection delays and detection stabilities are reported in Figures 3-4 in terms of the means and the standard deviations computed over the 50 randomly sampled onset times. The fault detection delays could be quantified precisely as the faults were induced at known times in terms of temperature trend onset times.

We found that the multi-target model could detect the induced faults at least as fast as the single-target model. In the majority of the fault instances in this case study, the multi-target MLP even enabled a somewhat faster detection of the gear bearing fault than the single-target network. As shown in Figures 3 and 4, the multi-target MLP facilitated shorter detection delays for both slow and fast trends and regardless of the chosen alarm criterion.

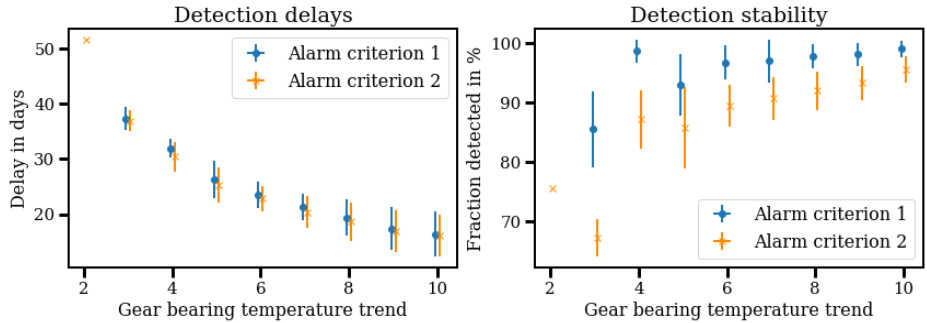

**Figure 3.** Fault detection delay and the detection stability are reported based on *multi-target* normal behavior MLP of the bearing temperature. The stability is the fraction of true positives from the first alarm until the end of the observation period, i.e. till the end of month 14. Mean and standard deviation of the detection delay and stability are reported for ten gear temperature trends. For each trend, the mean and standard deviations of detection delay and stability were computed from 50 randomly sampled temperature trend onset times in months 12 and 13. Mean detection delay and mean detection stability were computed based on randomized temperature trend onsets times and for 10 trend progression velocities, assuming a linear temperature trend. Fifty randomly selected onset times were used to simulate the bearing temperature increase.

This study demonstrates that multi-target models can describe the temperature behaviour of wind turbine gear bearings in normal operation at least as accurately as corresponding single-target models. In the present work, the multi-target MLP provided a test-set accuracy of 0.49 for predictions of the gear bearing temperature, while the test-set accuracy of the single-target MLP was 0.51 in normalized temperature. As the multi-target MLP produced smaller prediction residuals in many cases, temperature trends became visible earlier and thus the detection delay was shorter. In addition, a paired sample t-test was performed to test the null hypothesis that there are no systematic differences in the detection delays based on the normal behaviour descriptions of the multi-target MLP versus the single-target MLP. The null hypothesis was clearly rejected ($p < 10^{-21}$) in favor of the alternative hypothesis of shorter detection delays based on the multi-target MLP. The test result was confirmed for both alarm criteria.

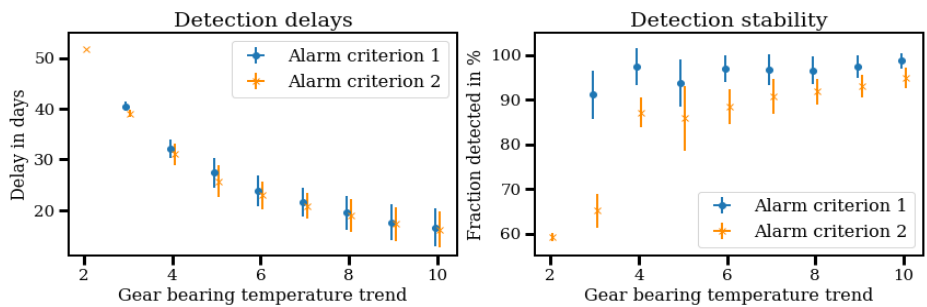

**Figure 4.** Fault detection delay and stability as in Figure 3 but based on the *single-target* normal operation MLP of the gear bearing temperature.

Similarly, a paired sample t-test was performed to test for systematic differences of the detection stability. This provided a more ambiguous picture. Based on alarm criterion 2, the multi-target MLP enabled a more stable alarm after the first detection ($p<10^{-14}$). However, no systematic difference in the detection stability was found in the case of alarm criterion 1 ($p=0.22$) Comparing both alarm criteria, it is also found that the second criterion detects faults significantly earlier than the first criterion, but this comes at the cost of significantly reduced detection stabilities (Figures 3-4).

In summary, this study demonstrated that multi-target MLPs can detect faults as fast as and in some cases even earlier than single-target MLPs, and at the same time achieve the same level of detection stability. Comparing Figures 3 and 4, the detection speed-up observed in this study ranged from several hours to several days. The earlier detection enabled by the proposed multi-target approach can deliver a significant advantage in the planning and performance of maintenance activities. If wind farm operators learn about a developing fault hours or up to several days ahead, they have more time to respond and schedule inspections and adjustment work that may prevent more serious damage and component replacement.

## 5    Conclusions

This study investigated how multi-target neural networks compare to single-target models with regard to the speed and accuracy of detecting incipient faults in wind turbines from SCADA data. We analyzed the delays in detecting gear bearing faults in onshore wind turbines. Gear bearing faults can result in anomalous temperature increases that are detectable from SCADA data. In this work, synthetic temperature trends were overlaid on the gear bearing temperature SCADA signals in order to facilitate a systematic analysis despite the scarcity of fault observations. We compared the detection delay and the detection stability of the alarm signals among the multi- and single-target models based on different alarm criteria. In this study, we found that multi-target neural networks can meet and even go below the detection delays achieved with single-target MLPs of the turbine normal operation. At the same time, the multi-target MLPs achieved the same level of prediction stability. We demonstrated that the detection of temperature-related faults could be accelerated by up to several days compared to the state-of-the-art fault detection with single-target models. With regard to future studies, we propose to also investigate the potential of multi-target models for normal behaviour modelling and fault detection tasks based on high-frequency data, in particular from vibration measurements in the drive train, which did not form part of this paper.

**Acknowledgments:** The author thanks Bernhard Brodbeck, Janine Maron, Dimitrios Anagnostos of WinJi AG, Switzerland, and Kaan Duran of Energie Baden-Wuerttemberg EnBW, Germany, for valuable discussions.